\documentclass[letterpaper, 10 pt, journal, twoside]{IEEEtran}

\usepackage{cite}
\usepackage{amsmath,amssymb,amsfonts}
\usepackage{algorithmic}
\usepackage{graphicx}
\usepackage{textcomp}
\usepackage{verbatim}
\usepackage[linesnumbered,ruled,vlined]{algorithm2e}
\usepackage{xcolor}
\usepackage{subcaption}
\usepackage{multirow}
\usepackage{amssymb}
\usepackage{pifont}
\usepackage{booktabs}

\newcommand{\cmark}{\ding{51}}%
\newcommand{\xmark}{\ding{55}}%

\begin{document}
%
\title{Grasping as Inference: Reactive Grasping in Heavily Cluttered Environment}

\author{Dongwon Son
\thanks{Manuscript received: January, 8, 2022; Revised April, 7, 2022; Accepted June, 1, 2022.}
\thanks{This paper was recommended for publication by Editor Markus Vincze upon evaluation of the Associate Editor and Reviewers' comments.}
\thanks{The author is with AI Method team, Samsung Research, Samsung Electronics, Seoul, Republic of Korea}%
\thanks{Digital Object Identifier (DOI): see top of this page.}
}


\markboth{IEEE Robotics and Automation Letters. Preprint Version. Accepted June, 2022}
{Dongwon Son: Grasping as Inference}

\maketitle

\begin{abstract}
Although, in the task of grasping via a data-driven method, closed-loop feedback and predicting 6 degrees of freedom (DoF) grasp rather than conventionally used 4DoF top-down grasp are demonstrated to improve performance individually, few systems have both. Moreover, the sequential property of that task is hardly dealt with, while the approaching motion necessarily generates a series of observations. Therefore, this paper synthesizes three approaches and suggests a closed-loop framework that can predict the 6DoF grasp in a heavily cluttered environment from continuously received vision observations. This can be realized by formulating the grasping problem as Hidden Markov Model and applying a particle filter to infer grasp. Additionally, we introduce a novel lightweight Convolutional Neural Network (CNN) model that evaluates and initializes grasp samples in real-time, making the particle filter process possible. The experiments, which are conducted on a real robot with a heavily cluttered environment, show that our framework not only quantitatively improves the grasping success rate significantly compared to the baseline algorithms, but also qualitatively reacts to a dynamic change in the environment and cleans up the table.
\end{abstract}


%
\IEEEpeerreviewmaketitle

\section{INTRODUCTION}
\begin{figure}
    \centering
    \includegraphics[width=0.49\textwidth]{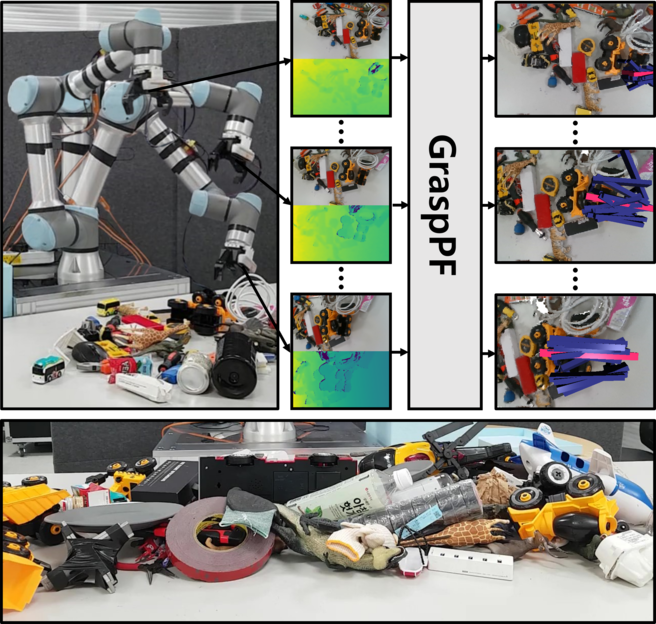}
    \caption{Grasp estimation flow (top) and heavily cluttered environment used in the experiment (bottom). The proposed framework can estimate the grasp in real-time by utilizing sequential observations, and finally, it can clean up the table.}\vspace{-3mm}
    \label{fig:heavily_cluttered_environment}
\end{figure}

This paper aims to find a robust solution for grasping objects in a heavily cluttered environment through partial RGB and depth (RGBD) vision observations which are obtained from a camera rigidly attached to the gripper, as shown in Fig. \ref{fig:heavily_cluttered_environment}. Recently, a lot of data-driven methods to address this problem have been presented. Specifically, grasping through predicting explicit grasp pose has been showing results. Whereas most of the research in the field focuses on improving the precision of the deep network, 
there are three other experimentally verified approaches which improve the success rate of the grasping task.
In the \cite{morrison2018closing}, they focus on the effects of the closed-loop grasping and verify that it improves not only robustness to control errors but also reactiveness to a dynamic environment. On the other hand, the grasp prediction models in \cite{murali20206, fang2020graspnet} use 6 degrees of freedom (DoF) grasp space rather than conventionally used 4DoF space (i.e., top-down), and they show this extension of grasp space is beneficial in the cluttered environment.
Concurrently, the work of \cite{morrison2019multi} finds that utilizing sequential multi-view information while approaching can improve performance. The three approaches are orthogonal to each other so that they can be synthesized together, but the problem is that few works achieve it as far as we know.

We hypothesize that merging benefits from three approaches can dramatically improve task success rate as well as generate reactive motion, and we verify it by developing a novel framework, which has the above-mentioned three properties, and a real robot experiment.
In short, we develop the closed-loop framework to estimate 6DoF grasp in a heavily cluttered environment, given the history of partial RGBD pixel input. We achieve it by 1) formulating the grasping problem as an online estimation problem and 2) developing a novel lightweight grasp prediction network.
We first formulate the problem as Hidden Markov Model (HMM) and derive a recurrent formulation to infer grasp.
This inference formulation is similar to Bayes filter\cite{thrun2005probabilistic}, then particle filter, which is one powerful realization of it, is applied. Through this framework, named Grasp Particle Filter (GraspPF), the robot can not only retain prior grasp distribution but also refine the grasps while approaching (i.e., online estimation).

However, to apply particle filter, a precise and lightweight grasp prediction model, which can generate grasps from the partial RGBD observation in real-time, and a grasp evaluation model, which can evaluate any grasp candidates in a bounded space, are required. Therefore we develop a unified data-driven model which can achieve both.
To realize this, we design Directional Grasp Quality CNN (DGQ-CNN), which predicts pixel-wise grasp quality from grasp rotation and RGBD inputs. The additional input of grasp rotation makes the network not restricted to a predefined rotation set, which is prevalent in prior works\cite{fang2020graspnet, wang2021graspness, ten2021efficient}. Additionally, it is computationally efficient enough to run in a closed-loop manner during the approach, making the resulting implementation reactive.
The proposed framework is verified with a comparison experiment with the state-of-the-art algorithms, and it outperforms them by a large margin in terms of success rate. Additionally, ablation studies show the performance improvement of individual factors.

In summary, the contribution of this paper is that, for the first time, we fuse three effective components in the data-driven grasp prediction field, which are 1) to utilize closed-loop feedback, 2) to predict 6DoF grasp, and 3) to utilize sequential multi-view observations. It can be achieved by introducing GraspPF and DGQ-CNN.

\begin{table}
\centering
\begin{tabular}{l c c c c c} 
\specialrule{1.0pt}{1pt}{1pt}
\multirow{2}{*}{Methods} & CL & Multi & \multirow{2}{*}{6DoF} & Offline & Dataset\\
& (Reactive) & View & & Training & Source \\
\hline
GG-CNN-cl\cite{morrison2018closing} & \cmark & \cmark & \xmark & \cmark & real \\
MVP\cite{morrison2019multi} & \xmark & \cmark & \xmark & \cmark & real \\
Dex-Net\cite{mahler2017dex} & \xmark & \xmark & \xmark & \cmark & syn \\
\cite{fang2020graspnet,wang2021graspness,li2021simultaneous} & \xmark & \xmark & \cmark & \cmark & real \\
\cite{ten2021efficient,mousavian20196,sundermeyer2021contact} & \xmark & \xmark & \cmark & \cmark & syn \\
GPD\cite{ten2017grasp} & \xmark & \cmark & \cmark & \cmark & real \\
QT-Opt\cite{kalashnikov2018qt} & \cmark & \xmark & \xmark & \xmark & real \\
Song et al. \cite{song2020grasping} & \cmark & \xmark & \cmark & \xmark & real \\
GraspPF (Ours) & \cmark & \cmark & \cmark & \cmark & both \\
\specialrule{1.0pt}{1pt}{1pt}
\end{tabular}
\caption{A comparison of our work to related approaches ("CL" stands for closed-loop.)}
\label{tab:realated_work}
\end{table}

\section{Related Works}
We compare the proposed algorithm with other works and summarise the results in Tab. \ref{tab:realated_work}.
The data-driven grasping method through predicting explicit grasp in the cluttered environment \cite{li2021simultaneous, wang2021graspness, sundermeyer2021contact, fang2020graspnet, ten2021efficient} usually follows the steps: 1) get the RGBD (or depth only) measurement at a global position which can view all objects in the fixed workspace 2) generate grasp candidates and choose one by utilizing a trained model 3) execute the open-loop pick and place motion with a target grasp configuration which is obtained in the previous step.
Specifically, the method in \cite{fang2020graspnet} follows the steps mentioned above with a grasp prediction model, which is trained with a real large-scale dataset, while \cite{sundermeyer2021contact} is with a synthetic dataset.
However, these methods do not sufficiently utilize the partial observation streaming during motions in step 3).

Although GG-CNN\cite{morrison2020learning} uses sequential observations with a light enough network to fulfill the closed-loop control, it does not sufficiently use the prior observations, because it just picks the closest grasp to the prior estimated grasp after local peak clustering. Additionally, the predicted grasp space of GG-CNN is restricted to 4DoF. 
On the other hand, the works of \cite{morrison2019multi, ten2017grasp} utilize multi-view observations to predict grasp pose in a cluttered scene by stacking predicted results in a discretized table \cite{morrison2019multi} or merging them into one point cloud \cite{ten2017grasp}. Still, both works are hard to generate reactive motion.
Compared with these, our approach can retain 6DoF grasp distribution as a form of particles and run on an online estimation manner, which enables it to react dynamic environment.

Another branch of the data-driven approach for grasping is directly predicting continuous action without estimating the explicit grasp. For example, works in \cite{kalashnikov2018qt, song2020grasping} realize closed-loop grasping through reinforcement learning. Still, they do not take advantage of past observations and, moreover, require real robot interaction, which makes the resulting model dependent on a particular robot.
On the other hand, in the grasp estimation methods, the large-scale open-source dataset \cite{eppner2021acronym, fang2020graspnet}, which is independent of the robot, can be used for training. Additionally, it can be trained offline without real robot interaction. For these reasons, we focus on the grasp estimation method while keeping the advantage of the closed-loop.

\begin{figure}
    \centering
    \includegraphics[width=0.45\textwidth]{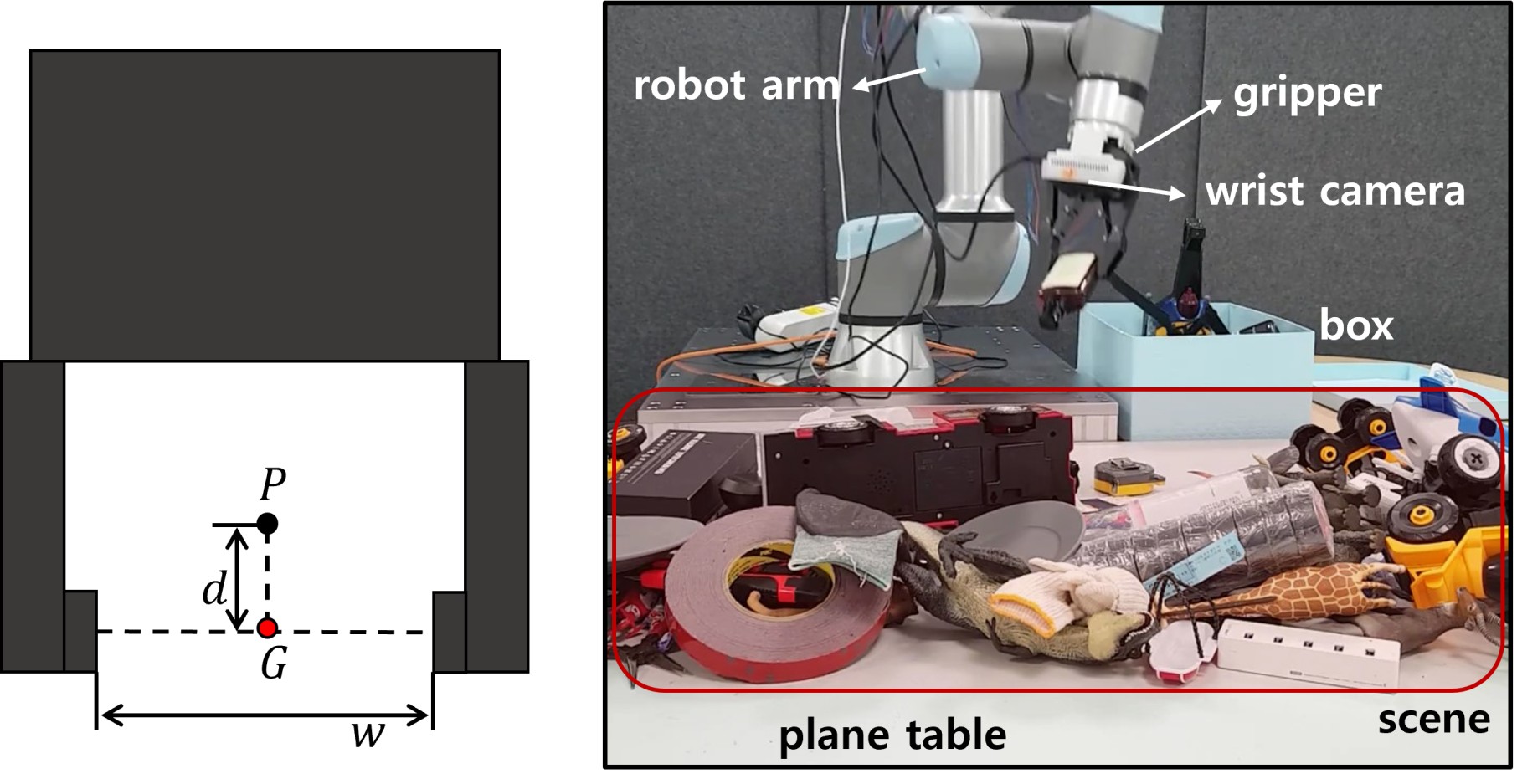}
    \caption{Illustration of grasp configuration (left) and setting of the grasping problem (right).}\vspace{-3mm}
    \label{fig:grasp_representation}
\end{figure}

\section{Problem Statement}
\label{sec:problem_statement}

In this paper, we assume that there are a scene and a robot arm, as shown in Fig. \ref{fig:grasp_representation}. The scene contains a flat table and cluttered objects which are rested on the table in a stable pose. Besides the scene, there is an N-DoF robot equipped with a two-finger parallel gripper on the end-flange, and a camera is rigidly attached to the gripper (i.e., wrist camera). 
The observation $z$ contains the pose of the camera $T^W_C\in SE(3)$ where $W$ denotes the world frame, the intrinsic parameter of the camera $\xi\in\mathbb{R}^6$, the RGB and depth images captured from the wrist camera.
Regarding grasp representation, the grasp configuration space is defined as $SE(3)\times\mathbb{R}\times\mathbb{R}$, which is also illustrated in Fig. \ref{fig:grasp_representation}. It includes the 6DoF pose $P \in SE(3)$ which consists of a point $p^W_P\in\mathbb{R}^3$ on the observed point cloud and rotation $R^W_P\in SO(3)$, gripper width $w\in\mathbb{R}$, and grasp depth $d\in\mathbb{R}$ of which direction is based on the gripper approaching axis (z-axis of $R^W_P$).
Then, grasp pose $G\in SE(3)$ can be calculated from $P$ and $d$.
Grasp width $w$ is discretized to 2 bins to simplify grasp space.

\begin{figure*}
        \begin{subfigure}{0.33\textwidth}
        \centering
        \includegraphics[width=\textwidth]{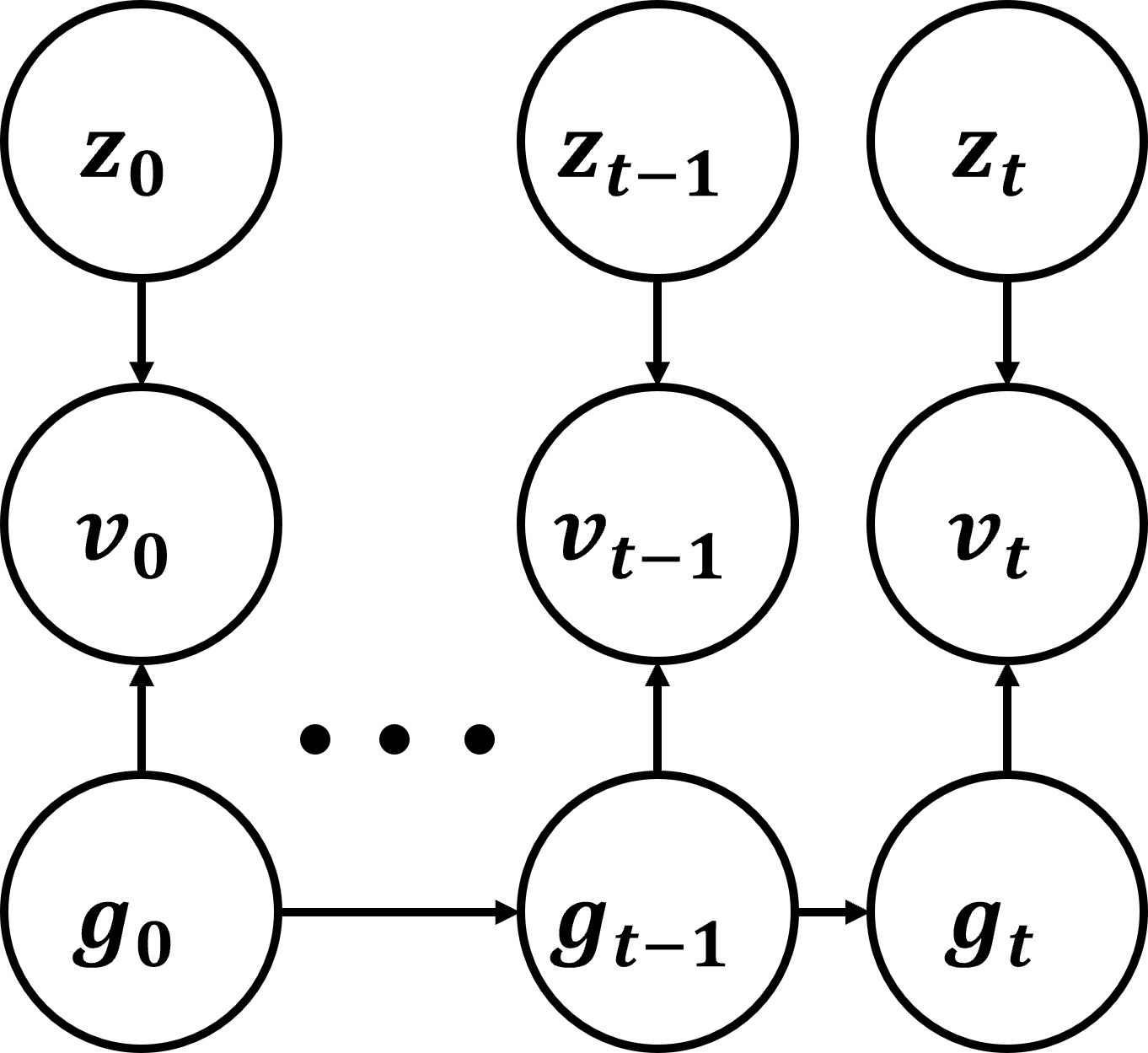}
        \caption{}
        \label{fig:bayes_graph}
        \end{subfigure}
        \hfill
        \begin{subfigure}{0.63\textwidth}
        \centering
        \includegraphics[width=\textwidth]{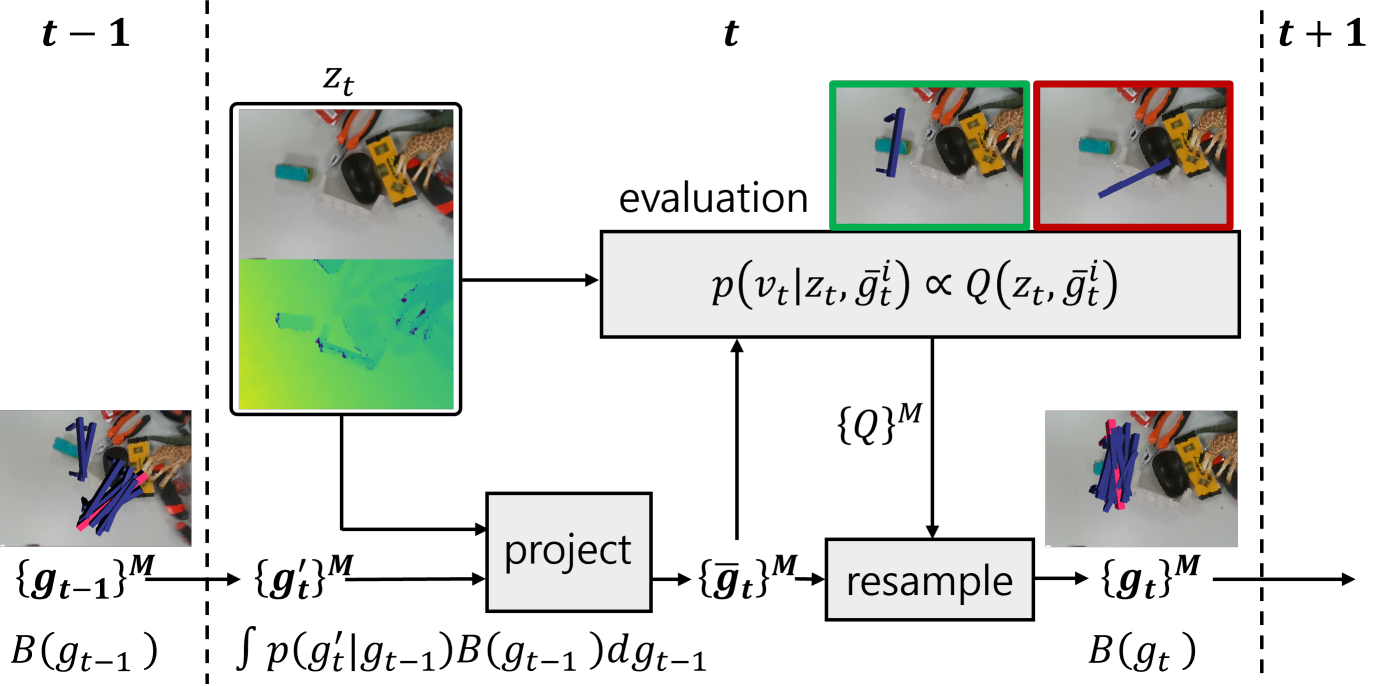}
        \caption{}
        \label{fig:GraspPF_sequence}
        \end{subfigure}
    \caption{Graphical model for HMM of grasping (a) and illustration of GraspPF (b). 
    The grasp particles $\{g_{t-1}\}^M$ undergo three steps to be propagated to $\{g_t\}^M$.
    First, they are transited by the noise model to be $\{g'_t\}^M$, then projected by observed points to be $\{\bar{g}_t\}^M$, and finally resampled with evaluation values of $\{Q\}^M$ to become $\{g_t\}^M$. 
    }\vspace{-3mm}
    \label{fig:GraspPF}
\end{figure*}

A successful grasp is defined as the configuration of grasping which has a high probability of succeeding in picking and placing the object without any collision between the gripper, robot arm, objects, and table. The grasps are evaluated by grasp quality measure $Q:SE(3) \rightarrow{} \mathbb{R}^2$, which predicts the quality value per each width bin from $G$.
Note that the symbol $Q$ is also used for the quality value itself in this paper. Finally, the problem this paper deals with is that, given a scene, a robot arm, and a wrist camera, the robot tidies up objects in the scene via estimating successful grasp distribution from the streaming of $z$ for each time step. 
In the next section, we formally express this problem and explain the connection with the online estimation.

\section{Grasping as Inference}
\label{sec:grasping_as_inference}

\subsection{HMM and Online Estimation for Grasping}

A grasping problem can be expressed as estimating current grasp distribution conditioned on the history of observations as $p(g_t|z_{0:t}, v_{0:t}=1)$ where $\cdot_{a:b}$ is history of variables from time step $a$ to $b$, $g_t$ is the grasp as defined in Sec. \ref{sec:problem_statement}, $z_{0:t}$ is history of observation from time step 0 to t, and $v_{0:t}$ is binary success variable which is 1 if task is success and 0 otherwise, which is also introduced in \cite{mahler2017dex}. Note that $p(\cdot|v=1)$ is simplified to $p(\cdot | v)$, because we do not consider cases conditioning on failure. In this paper, the grasp estimation problem is formulated as an inference in HMM as shown in Fig. \ref{fig:bayes_graph}. We assume that $g_t$ is dependent on prior grasp $g_{t-1}$ because we can reasonably assume there is an optimal grasp distribution of the scene in terms of the grasp success metric, so the current belief state of the grasp is closely related to the prior one. And we set grasp success variable $v_t$ dependent to $g_t$ and $z_t$.
Then, the belief state of $g_t$, defined as $B(g_t)=p(g_t|v_{0:t},z_{0:t})$, can be propagated from prior time step as,
\begin{equation}
    B(g_t) = \mu p(v_t|g_t,z_t) \int p(g_t|g_{t-1})B(g_{t-1}) \, dg_{t-1}
    \label{eq:bayes_filter}
\end{equation}
where $\mu$ is normalization factor.
This formulation has a recurrent structure and it is similar to Bayes filter except that the motion model (or transition model) $p(g_{t}|g_{t-1})$ does not involves action and measurement model is expressed as $p(v_{t}|g_t,z_t)$ by introducing $v$. That means well-established tools for estimation problems can be utilized to solve this. 
To apply estimation toolset, motion model $p(g_t|g_{t-1})$ and measurement model $p(v_t|g_t,z_t)$ should be identified first.
We reasonably assume that $p(g_t|g_{t-1})$ follows Gaussian distribution as noise model because the scene can be regarded as temporarily static between time step $t-1$ and $t$. On the other hand, measurement model $p(v_t|g_t,z_t)$
can be obtained by calculating $Q$.
In the next section, the realization method of recurrently solving \eqref{eq:bayes_filter} is explained.

\subsection{Grasp Particle Filter}
\label{sec:GraspPF}
\begin{algorithm}[t!]
\SetAlgoLined
\KwResult{grasp estimation result}
recieve initial observation $z$\\
$\{g\}^M=GetInitialDistribution(z)$ \\
 \For{each time step}{
 $\{g'\}^M=Transition(\{g\}^M)$\\
 receive observation $z$\\
 $\{\bar{g}\}^M=Projection(z,\{g'\}^M)$ \\
 $\{Q\}^M=Evaluation(z, \{\bar{g}\}^M)$\\
 $\{g\}^M \leftarrow Resampling(\{\bar{g}\}^M, \{Q\}^M)$
 }
 \caption{GraspPF}
 \label{algorithm:GraspPF}
\end{algorithm}

\begin{figure*}
    \centering
    \includegraphics[width=0.97\textwidth]{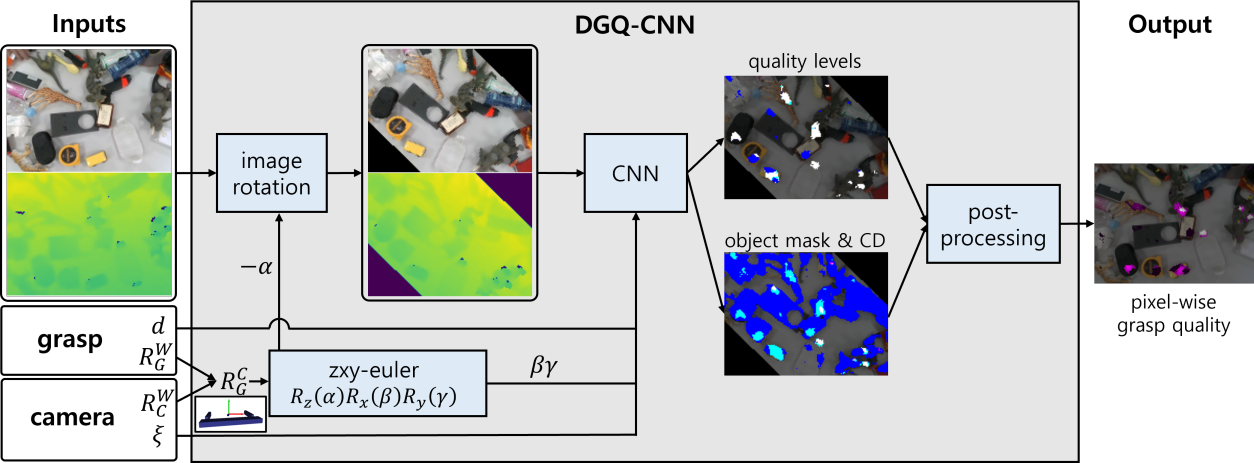}
    \caption{Illustration of DGQ-CNN. DGQ-CNN includes pre-processing, parameterized CNN, and post-processing. DGQ-CNN predicts pixel-wise grasp quality by receiving observed RGBD image, grasp rotation $R^W_G$, grasp depth $d$, camera pose $R^W_C$, and camera intrinsic parameter $\xi$. The pre-processing includes zxy-Euler parameterization of $R^C_G$, and image rotation. And the post-processing contains extraction of continuous grasp quality values from 6 channels of prediction which consist of 3 levels of grasp quality, object mask, and collision detection mask per two widths (CD in the figure).}\vspace{-3mm}
    \label{fig:DGQ-CNN}
\end{figure*}

As explained before, grasp estimation can be formulated as Bayes filter, but it includes highly non-linear models, making the realization not trivial. Among the possible realizations, we use particle filter\cite{thrun2005probabilistic}, because it has been verified to work well with non-linear model\cite{deng2021poserbpf, manuelli2016localizing, wirnshofer2019state}. 
The resulting algorithm is GraspPF, which is illustrated in Fig. \ref{fig:GraspPF_sequence} and Alg. \ref{algorithm:GraspPF}, and a detailed description follows. The first step is the initialization of the successful grasp distribution. 
The belief state of grasp $B(g_{t-1})$ is approximated to $M$ particles in grasp configuration space $\{g_{t-1}\}^M$ defined in Sec. \ref{sec:problem_statement}, where $\{\cdot\}^X$ means batch of size $X$. Then time update is applied to these particles with $\int p(g_{t}|g_{t-1})B(g_{t-1})dg_t$, resulting in $\{g'_{t-1}\}^M$. This transition is easily implemented by noise sampling for each particle. Specifically, we apply Gaussian noise to each $d$, $p^W_P$, and $R^W_P$. However, the transition through noise distribution makes the particles violate a constraint, where $p^W_P$ should be on the surface of the scene. To resolve it, we project each $p^W_G$ of particles to the scene mesh based on current depth observation included in $z_{t}$. Concretely, in the grasp space, the $p^W_{P,i}$ of the particle is projected to $\bar{p}^W_{P,i}$, which is the observed points following line of sight, then the batch of particles becomes $\{\bar{g}_t\}^M$. 
Afterward, the particles are evaluated by the approximated grasp quality measure function $Q$, which is conditioned on the current observation $z_{t}$. Then the current belief state $B(g_{t})$ is obtained from the resampling based on the evaluation values $\{Q\}^M$. These procedures are repeated during the approaching motion, and when the target is close enough, the scripted closing motion is executed. 

Throughout the GraspPF process, the past information is kept in prior distribution $B(g_{t-1})$ in the form of particles, and they are refined with continuously received observations. It makes GraspPF entertain sequential information while the measurement model only sees current observation. 
It is also worth noting that this process can be used as iterative refinement with a fixed observation (i.e. open-loop), which can resolve the roughness of the initial distribution. We demonstrate it in the comparison experiments on the real robot as explained in Sec. \ref{sec:experiment_on_real_robot}.
However, to get this inference framework to work, there are two unresolved problems: evaluation and initialization. Let us explain each problem and what makes them challenging.

First, the evaluation problem can be regarded as obtaining $Q$. To obtain exact results, the oracle information of scene and gripper is needed, such as the mesh and pose of each object. Unfortunately, this information is hard to be accessible with real robot settings, so many works adopt bypass methods to predict it with implicit information from the RGBD pixel. This can be achieved by learning-free methods such as \cite{adjigble2018model}, but these are hard to be extended to partial depth observation and are also challenging to be computationally efficient, which can be achieved by data-driven methods\cite{mahler2017dex, mousavian20196, ten2017grasp}.

On the other hand, the reasonable initial grasp distribution is also essential to the performance since approaching motion takes only a few seconds, limiting the number of iterations in GraspPF to relatively small number. However, the initialization is not trivial because grasp space is continuous, and the result should sufficiently cover the workspace.
In this paper, we resolve these evaluation and initialization problems simultaneously by introducing a novel network design, which can not only evaluate grasp candidates but also generate them efficiently, whereas authors in \cite{mousavian20196,mahler2017dex} introduce additional network or antipodal heuristics for sampling. In the next section, we explain the details of the network.

\subsection{Directional Grasp Quality Network}
\label{sec:directional_grasp_quality_network}

Before starting a detailed explanation of the network, we sum up the necessary properties for the network model in GraspPF, which are mentioned in the previous section.
First, it should be computationally efficient enough for a closed-loop to refine grasp during approaching motion. Second, it should be able to evaluate any grasp in a bounded space, which is necessary for the measurement model in GraspPF.
Lastly, it should be able to generate grasp candidates efficiently to get a reasonable initial distribution of grasps. We found that these can be achieved by a novel concept of the directional grasp quality.

Directional grasp quality is defined as grasp quality conditioned on fixed grasp rotation $R^W_G$. 
To utilize this concept, we design DGQ-CNN to predict the pixel-wise directional grasp quality from input as RGBD image and fixed rotation $R^W_G$ as Fig. \ref{fig:DGQ-CNN}. We first express grasp rotation as zxy-Euler parameters as $R^C_G=R_z(\alpha)R_x(\beta)R_y(\gamma)$, where $R^C_G$ is rotation matrix of $G$ with respect to camera frame $C$, $R_z(\alpha)$ is rotation matrix through an angle $\alpha$ with respect to $x$ axis, and so does $R_x(\beta)$ and $R_y(\gamma)$. Then, $\beta, \gamma$ are included in the inputs of the parameterized CNN, and the $\alpha$ is applied to the RGBD image, which is rotated by $-\alpha$ inspired by \cite{zeng2018robotic}. Additionally, the grasp configuration also contains continuous grasp depth $d$ and grasp width $w$, and we set $d$ as an additional input of the CNN and make $w$ as output with the discretization of size 2. In the perspective of CNN, it receives the rotated images and $\beta, \gamma, d, \xi$, and predicts pixel-wise grasp quality for 3 levels, object mask, and valid collision mask per two grasp width $w$, and then total output channel is 6 (3+1+2). Finally, the pixel-wise directional grasp quality is calculated by network outputs after post-processing, which contains extracting the continuous final grasp quality values and Gaussian filter.
It is worth specifying that the DGQ-CNN lessens the sampling burden by deleting $p^W_P$ in the sampling space. To be specific, DGQ-CNN receives only the rotation component $R^W_P$ in $P$ except $p^W_P$, and $p^W_P$ can be reconstructed by the point cloud and pixel index operation.

Once DGQ-CNN is obtained, the sampling and evaluation are straightforward.
In the sampling, the rotation $R^W_G$ and grasp depth $d$ are sampled uniformly within the predefined bound, and they enter into DGQ-CNN, which generates the pixel-wise quality prediction $Q$. Then the grasp candidates are recovered by pixel indices where $Q$ is over the threshold. Finally, the grasps are sampled proportionally in terms of grasp quality $Q$. 
On the other hand, regarding the evaluation, $Q$ can be obtained from the index-gathering after pixel-wise grasp quality prediction by DGQ-CNN with input as grasp configuration and observation $z$. Therefore introducing DGQ-CNN resolves the initialization and evaluation issue simultaneously. Next, details about the dataset and training for DGQ-CNN are explained.

\subsection{Dataset and Training}
\label{sec:dataset}

\begin{figure}
    \centering
    \includegraphics[width=0.45\textwidth]{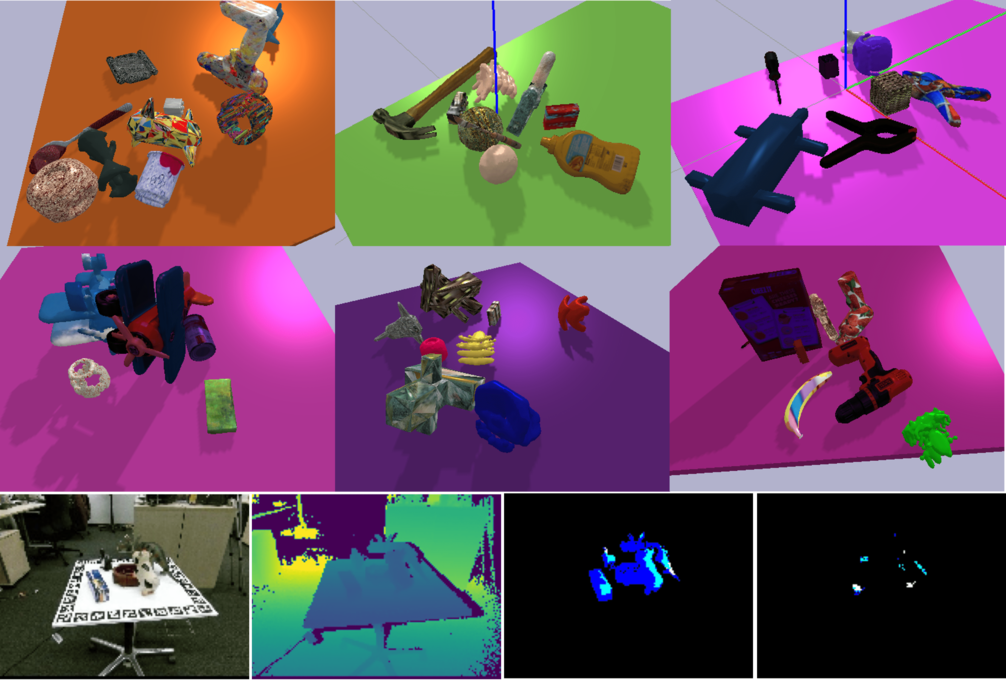}
    \caption{Synthetic scenes used to generate dataset (first two rows), and real RGB and depth image with corresponding labels within the dataset (last row). Texture, scale, and color are randomized in the synthetic scene. Two RGB images render the label. The 3 channel values of the first image are binary values of object mask and collision results per 2 widths, and the masks for the second image are binary values of 3 levels of grasp quality.}\vspace{-3mm}
    \label{fig:synthetic_real_dataset}
\end{figure}

To generate the dataset for training DGQ-CNN, a labeling method is needed to measure grasp quality (or how successful the grasp is). With the results from \cite{mahler2017dex, sundermeyer2021contact, mousavian20196}, we utilize synthetic labeling because directional grasp quality per pixel needs massive evaluations, and it is not accessible to real grasp\cite{song2020grasping} or human labeling\cite{zeng2018robotic}. 
We use the force-closure based metric motivated by \cite{ten2017grasp,mahler2017dex,fang2020graspnet}, where a prediction models trained on that metric are verified to perform well, and particularly a validation experiment with real grasping is conducted in \cite{fang2020graspnet}. Although some works such as \cite{kappler2015leveraging} report opposite results, we find that it also works well in our case. Specifically, grasp quality for each pixel has 3 dimensions, which correspond with 3 levels of grasp quality. Also, the label contains object mask and valid collision results per width evaluated by the simplified gripper in Fig. \ref{fig:grasp_representation}, and then the final labels have 6 channels as shown in Fig. \ref{fig:synthetic_real_dataset}.

The labeling method above can be applied to a real dataset where 3D mesh and RGBD observation are carefully matched, for example, \cite{kaskman2019homebreweddb,xiang2017posecnn}, and this approach is used in \cite{ten2017grasp, fang2020graspnet}. However, the sole source of the real dataset is insufficient to train DGQ-CNN because it needs vision data at various distances to work well even at close proximity to objects. So we generate additional 
data from various distances to the scene and camera parameters by utilizing synthetic rendering and simulation\cite{coumans2019}.
As previously studied in \cite{eppner2021acronym}, various object set is also important, so we use multiple sources\cite{morrison2020egad, kaskman2019homebreweddb, calli2015ycb}.
A total of 2.1M synthetic and 400K real RGBD pairs are used.

After generating the dataset, the parameterized CNN in DGQ-CNN can be trained with binary classification loss. However, the dataset has a critical problem, sparse segmentation. It means that the positive pixels in the pixel space are too sparse, then the naive binary cross-entropy loss causes the well-known pain, reducing training stability and performance \cite{jadon2020survey}. 
Therefore we solve it by utilizing the hierarchical structure of the label. 
We decompose probability of grasp success, conditioned on pixel index and additional inputs of CNN as $\Theta$, into the three conditional probabilities as,
\begin{align*}
    p(success|\delta_{ij},\Theta) \propto& p(v,m_c,m_o|\delta_{ij},\Theta)\\
    =&p(v|m_c,m_o,\delta_{ij},\Theta)p(m_c|m_o,\delta_{ij},\Theta)\\
    &p(m_o|\delta_{ij},\Theta)
\end{align*}
where $\delta_{ij}$ is pixel in index $i$ and $j$, $m_o$ is binary random variable of object presence, and $m_c$ is binary random variable for valid collision. Then we make DGQ-CNN predict each conditional probability $p(v|m_c,m_o,\delta_{ij},\Theta)$, $p(m_c|m_o,\delta_{ij},\Theta)$ and $p(m_o|\delta_{ij},\Theta)$, and this decomposition reduce imbalance problem significantly.

\section{Experiment}
The final step of this paper is to apply GraspPF on a real robot. We use 6 DoF UR5e robot equipped with Robotis Hand (RH-P12-RN), CPU of Intel i9-9900K with 3.60GHz, and GPU of Nvidia RTX 4000 as verification settings. The object set contains tools, dishes, cups, and toys, as shown in Fig. \ref{fig:heavily_cluttered_environment}, that are prevalent in the house and not contained in the dataset. To make a real robot perform the task, the controller and task planning are needed in addition to the grasp estimation framework. This section explains the implementation details, protocol, and results of the experiments.

\subsection{Controller Structure on Real Robot}

\begin{figure}
    \centering
    \includegraphics[width=0.48\textwidth]{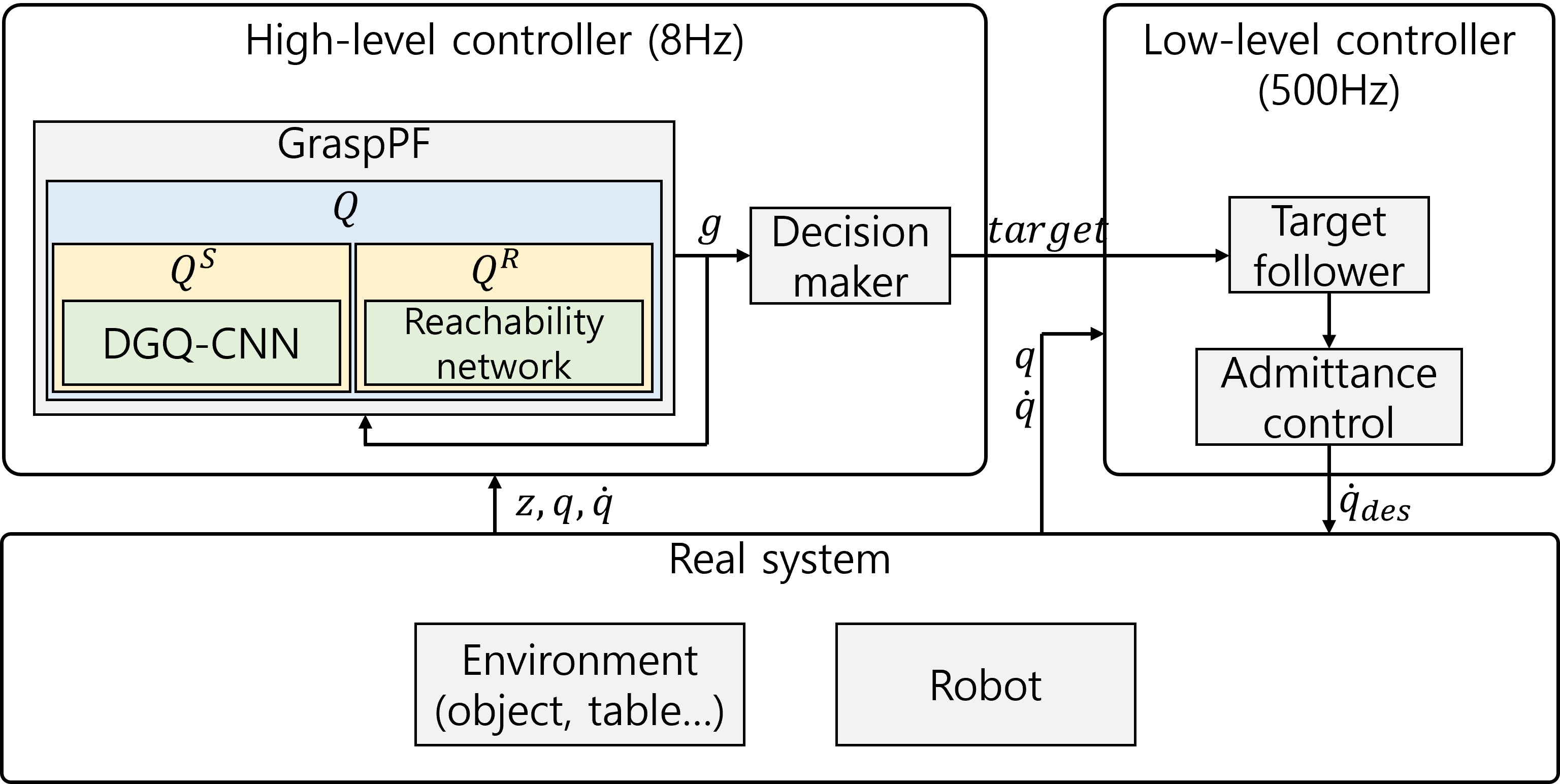}
    \caption{Controller structure of the experiment. The high-level controller runs on 8Hz and contains modules of GraspPF and decision-maker. GraspPF continuously estimates grasp from observation $z$ and sends it to the decision-maker module. And the decision-maker decides on high-level task semantics such as move-to-box or move-to-target. The low-level controller, which runs on 500Hz, receives the target from the high-level controller, joint state $q$, and joint velocity $\dot{q}$, and calculates desired joint velocity $\dot{q}_{des}$ in the configuration space of the robot.
    }\vspace{-3mm}
    \label{fig:real_robot_block_diagram}
\end{figure}

In the experiment, we set two layers of hierarchy as Fig. \ref{fig:real_robot_block_diagram}: high-level controller and low-level controller. First, the high-level controller runs on 8Hz, which contains GraspPF to estimate successful grasp, and the decision-maker, which decides on high-level motion semantics such as move-to-target or move-to-box. And the low-level controller includes target follower and admittance control. The simplest way to make a gripper approach to the grasp target is by calculating the vector from the gripper to the grasp target in the workspace and moving the robot arm in that direction. However, it does not consider the gripper approach direction and causes a collision between the gripper tip and the object. To prevent that, we design a specific target following controller considering the approaching direction. It is based on the waypoints, and the details are skipped because it is out of the scope of this paper.

\subsection{Reachability Network}
In the real robot implementation with the proposed framework and heavily cluttered environment, there are cases where the robot cannot reach the target due to kinematic constraint or self-collision. To resolve this reachability issue, we split $Q$ into two sub-functions, grasp quality measure for robot $Q^R$ and scene $Q^S$. $Q^S$ evaluates grasps based on collision between scene and gripper and grasp quality with a target object, which can be predicted by DGQ-CNN as explained in Sec. \ref{sec:directional_grasp_quality_network}.

In the case of $Q^R$, which evaluates grasps in terms of kinematics and self-collision, it is relatively easy to obtain because the calculation of reachability needs the occupancy and configuration of the robot, which can be accessible precisely even in the real robot.
Therefore, it can be calculated without any approximation, but empirically the methods based on collision-check and inverse kinematics (IK) are not feasible to check all particles $\{g\}^M$ in real-time when $M$ is large. Therefore, we introduce a simple neural network to approximate $Q^R$, in which data is generated by simulator\cite{coumans2019}. Regarding training details, the states in the workspace are sampled uniformly within bound, and then they are labeled based on IK and self-collision results. Afterward, the simple 2 layers Multilayer Perceptron (MLP) is trained with binary classification loss. The final reachability network only takes 100 micro-seconds to check 1500 grasp candidates, and its accuracy is 97\%.


\subsection{Experiment Results}
\label{sec:experiment_on_real_robot}

\begin{figure*}
    \centering
    \includegraphics[width=0.98\textwidth]{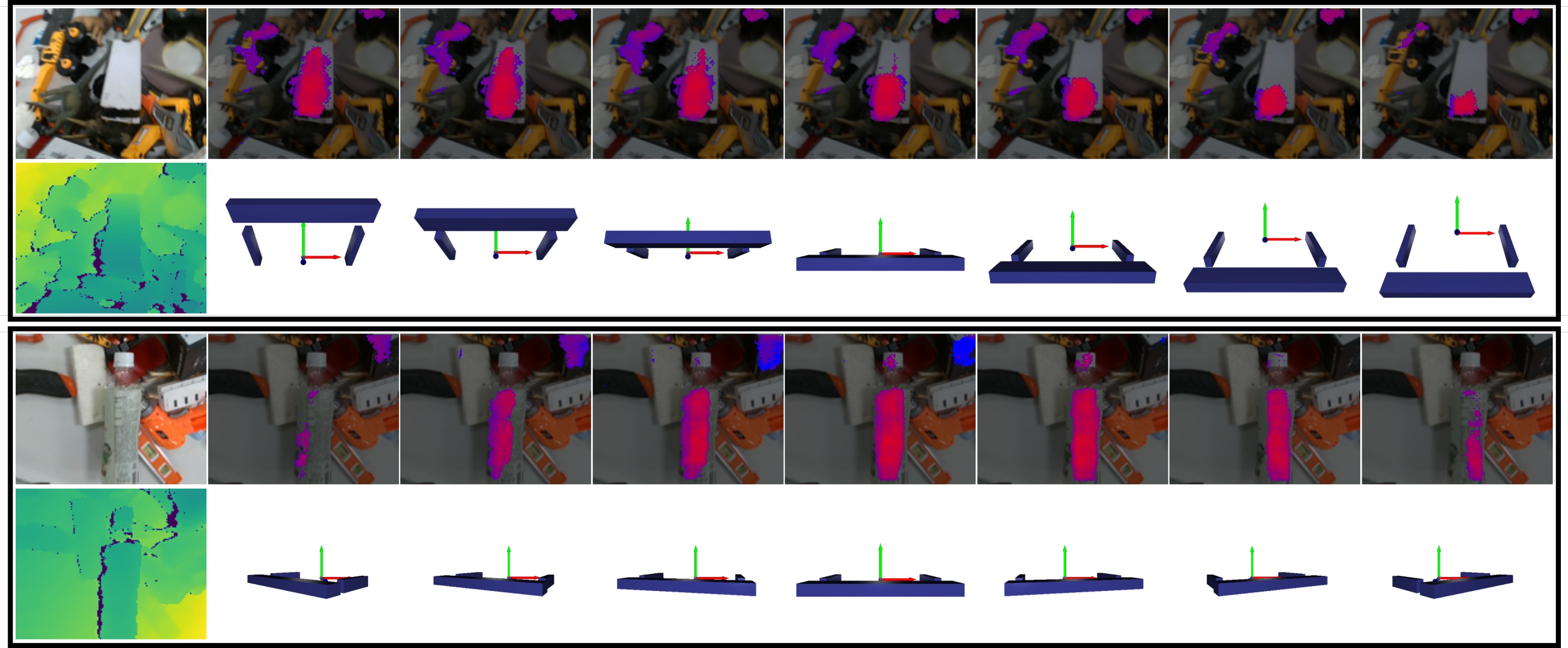}
    \caption{Visualization of prediction results according to grasp direction change. RGB and depth images as observations are shown in the first column, and among 4 rows, the first and third rows are predicted grasp quality results, where the higher the score, the redder it is. The second and fourth rows are visualization of rotated grasp, which is used for prediction. }\vspace{-3mm}
    \label{fig:pred_per_direction}
\end{figure*}

In this section, we present experimental results to quantitatively and qualitatively verify the performance of GraspPF and DGQ-CNN.
First, to compare the prediction performance of DGQ-CNN with our grasp quality intuition, the grasp quality prediction results per direction are visualized in Fig. \ref{fig:pred_per_direction}. In the prediction example of the first row, a box-shaped object is placed at an angle. Then, the high-quality region at the grasp direction, which matches the inclined angle, has a larger area than in other directions (second column). Similarly, in the second example, the high-quality region varies from left to right following the transition of the y-Euler angle. These examples show that additional grasp rotation input can change high-quality areas matching well with our intuition.

To verify GraspPF quantitatively, a comparison experiment is conducted with baseline algorithms to predict grasp. We adopt GraspNet\cite{fang2020graspnet}, Contact-GraspNet (abbreviated to Con-GraspNet)\cite{sundermeyer2021contact}, and GG-CNN\cite{morrison2018closing} as baselines because they are reported to show good performance in a cluttered environment, and the original implementations are provided. We use the pre-trained models provided by authors despite using a different gripper model (Robotis Hand) in our experiment than that of baselines. 
We reduce the impact by utilizing the predicted grasp width for pre-grasp motion, making the interaction between the gripper and other entities minimal.
GG-CNN is executed in a closed-loop manner, whose implementation reproduces the original Kinova Mico experiment.
Throughout all baselines, we filter out grasp candidates by $Q^R$.
GraspPF is executed in two versions: GraspPF-ol, GraspPF-cl. GraspPF-ol is an open-loop version of GraspPF, which uses the particle filter as a refinement with a fixed observation. On the other hand, GraspPF-cl runs in a closed-loop manner and continuously refines grasps during approaching motion.

Additionally, we also implement five ablation versions of GraspPF: Sampling-ol, Sampling-cl, GraspPF-TD, GraspPF-real, and GraspPF-syn.
Sampling-ol is a method that predicts grasps through forward sampling by DGQ-CNN and $Q^R$ (without refinement) from global observations at a fixed camera pose, where grasping is executed in an open-loop manner. 
Sampling-cl is a closed-loop version of Sampling-ol, where a batch of grasps is predicted every time step, and the grasp is selected by predicted quality and distance regularizer from the previously selected grasp. 
GraspPF-TD is the same as GraspPF-cl except for reducing grasp space to top-down (i.e., 4DoF). Lastly, in GraspPF-real, only real data is used to train DGQ-CNN, whereas GraspPF-syn uses only synthetic data.

The experiment protocol is as follow: 1) 62 unknown objects are placed in the box 2) pour them on the flat table to construct a heavily cluttered environment 3) set the initial gripper pose with a variety of distance to the table which is uniformly selected within 0.45-0.65m 4) execute the grasping algorithm and log the result 5) repeat 3-4) for 20 times and check the success rate. We repeat the experiment 7 times for each algorithm, and the total number of grasping trials is 140.
The comparison results are reported in the Tab. \ref{tab:experiment_result}. 

\begin{table}[h!]
\centering
\begin{tabular}{|l ||c c c c c c c c|} 
\hline
\multirow{2}{*}{Methods} & \multicolumn{8}{c|}{scene reset trial no. (\#/20 per each trial)} \\
& 1 & 2 & 3 & 4 & 5 & 6 & 7 & total\\
\hline
\textbf{Baselines} & & & & & & & &\\
GraspNet\cite{fang2020graspnet} & 13 & 13 & 16 & 14 & 14 & 15 & 12 & 69\% \\
Con-GraspNet\cite{sundermeyer2021contact} & 12 & 13 & 11 & 13 & 15 & 13 & 13 & 64\% \\
GGCNN-cl\cite{morrison2020learning} & 9 & 9 & 7 & 12 & 9 & 10 & 10 & 47\% \\
\hline
\textbf{Proposed} & & & & & & & &\\
GraspPF-cl & 16 & 17 & 16 & 18 & 17 & 19 & 17 & \textbf{86\%} \\
GraspPF-ol & 14 & 15 & 14 & 17 & 16 & 16 & 15 & 76\% \\
\hline
\textbf{Ablations} & & & & & & & &\\
Sampling-cl & 14 & 12 & 14 & 18 & 12 & 13 & 14 & 69\% \\
Sampling-ol & 15 & 12 & 11 & 12 & 14 & 15 & 11 & 64\% \\
GraspPF-TD & 11 & 13 & 14 & 19 & 18 & 13 & 17 & 75\% \\
GraspPF-real & 7 & 7 & 9 & 6 & 7 & 9 & 7 & 37\% \\
GraspPF-syn & 14 & 18 & 16 & 15 & 17 & 14 & 14 & 77\% \\
\hline
\end{tabular}
\caption{Results of the comparison experiment}
\label{tab:experiment_result}
\end{table}
First, in terms of open-loop performance, the proposed algorithm GraspPF-ol outperforms baseline algorithms GraspNet and Con-GraspNet. Furthermore, the performance margin increases to 17\% when applying closed-loop (GraspPF-cl). On the other hand, GraspPF-cl exceeds GGCNN-cl by 39\%. We find that GGCNNs struggle to predict the quality grasp in a highly dense cluttered environment. Next, to study the individual influence of each factor, the results of 5 ablation experiments can be compared. Fixing approach direction to top-down (GraspPF-TD) degrade performance when compared with GraspPF-cl, and this result can be evidence that introducing 6DoF grasp space is beneficial, which is also studied in \cite{li2021simultaneous}. Next, to see how introducing a closed-loop affects performance, GraspPF-cl, Sampling-cl, GraspPF-ol, and Sampling-ol can be compared. Overall, applying closed-loop brings performance gain, which comes from improved robustness to control errors and sensor noise as previously studied in \cite{morrison2020learning}.

Regarding the influence of introducing particle filter, which means extracting more information from sequential observations, the success rate of Sampling-cl shows a 16\% performance drop. This result implies that the system can keep the prior distribution better through particle filter and update the distribution with multi-view observations.
On the other hand, the effect of introducing particle filter as a refinement step can be revealed by comparing Sampling-ol and GraspPF-ol, and it shows an improvement of 13\%. This improvement can be explained by the fact that forward sampling is based on Monte-Carlo sampling, so it is not enough to cover a large continuous grasp space, although the space is reduced from pixel-wise prediction as explained in Sec. \ref{sec:directional_grasp_quality_network}. Then particle filter can refine the initial grasps and find better grasps even with the same observation. Lastly, regarding the effects of dataset source, although performance declines in both GraspPF-real and GraspPF-syn, it degrades much more in GraspPF-real. A slump in GraspPF-real is because of the lack of variation of camera parameters, including intrinsic and viewpoint, while GraspPF needs more dynamic variations. The 9\% drop in GraspPF-syn can be mainly from a sim-to-real gap in the depth sensor.

Furthermore, in the real cleaning-up experiment, the GraspPF-cl can chase an object, re-grasp after fail, and react to a sudden object change. These make the framework can clean all objects containing flat or deformable objects. The grasping failures mainly result from slippage. Specifically, the slippage arises while gripping because the estimated grasp region has a too obtuse angle, or the axial slippage results in the object's fall during the placing because of not considering the center of gravity.

\section{conclusion}
In this paper, we show that significant performance improvement can be obtained by merging three effective but orthogonal approaches: running as a closed-loop, extending grasp space to 6DoF, and utilizing sequential multi-view observations. They can be achieved by developing two main components, which are 1) GraspPF, enabling the framework to retain reasonable prior grasp distribution as a form of particles and update grasps from continuously received observations, and 2) DGQ-CNN, which is computationally efficient enough to run on a closed-loop in test time and unleashes restriction of grasp rotation to fixed set by setting direction information as input of the network. The resulting framework outperforms the state-of-the-art algorithm on the real robot experiment in terms of success rate in the heavily cluttered environment. Also, qualitatively it can chase the object by tracking the grasp, and react to a sudden object change.

Future work is an extension to more complex tasks by utilizing grasping as an action primitive. This includes discovering unintended behavior or the long horizon task.
Additionally, the high cost is one of the main discouragement of the household robot, and grasping via a cost-efficient robot is an interesting direction. The cost can be saved by reducing the DoF of the robot, and fortunately, DGQ-CNN can generate various grasps, which can be filtered out by an adequate feasibility model.

\section*{Acknowledgment}

The authors would like to thank Myungsin Kim, Unkyu Park, Jieun Park, Jaecheol Sim, Wonsik Shin, Joonmo Ahn, Jeongmin Lee, Jaeyoung Lim, Jaemin Yoon, Jaesik Chang, Rakjoon Chung, and Daekyoung Jung for their help on reviewing the idea and manuscript.

\bibliographystyle{IEEEtran}
\bibliography{IEEEabrv, ref}

\end{document}